\documentclass{article}

\usepackage{arxiv}

\usepackage[utf8]{inputenc} % allow utf-8 input
\usepackage[T1]{fontenc}    % use 8-bit T1 fonts
\usepackage{hyperref}       % hyperlinks
\usepackage{url}            % simple URL typesetting
\usepackage{booktabs}       % professional-quality tables
\usepackage{amsfonts}       % blackboard math symbols
\usepackage{nicefrac}       % compact symbols for 1/2, etc.
\usepackage{microtype}      % microtypography
\usepackage{lipsum}
\usepackage{graphicx}
\usepackage{comment}
\usepackage{amsmath}
\usepackage{natbib}
\usepackage{listings}
\graphicspath{ {./images/} }

\title{Click it or Leave it: Detecting and Spoiling Clickbait with Informativeness Measures and Large Language Models}

\author{
Wojciech Michaluk \\
Faculty of Mathematics and Information Science\\
Warsaw University of Technology\\
\texttt{wojciech.michaluk.stud@pw.edu.pl}
\And
Tymoteusz Urban \\
Faculty of Mathematics and Information Science\\
Warsaw University of Technology\\
\texttt{tymoteusz.urban.stud@pw.edu.pl}
\And
Mateusz Kubita \\
Faculty of Mathematics and Information Science\\
Warsaw University of Technology\\
\texttt{mateusz.kubita.stud@pw.edu.pl}
\And
Soveatin Kuntur \\
Faculty of Mathematics and Information Science\\
Warsaw University of Technology\\
\texttt{soveatin.kuntur.dokt@pw.edu.pl}
\And
Anna Wróblewska \\
Faculty of Mathematics and Information Science\\
Warsaw University of Technology\\
\texttt{anna.wroblewska1@pw.edu.pl}
}

\begin{document}
\maketitle
\begin{abstract}
%Clickbait headlines degrade the quality of information and user trust on the internet. We present a hybrid approach to clickbait detection that combines transformer-based text embeddings with linguistically motivated informativeness features and our \emph{baitness} score that quantifies headline “clickbaitness.” Using natural language processing, we evaluate classic vectorizers, word-embedding baselines, and large language model embeddings paired with tree-based classifiers. Our best model -- XGBoost over embeddings plus 15 features -- achieves an F1-score of 91\%, outperforming TF–IDF, Word2Vec/GloVe, and feature-only baselines. The feature set supports transparent, confidence-calibrated predictions by highlighting cues such as second-person pronouns, superlatives, numerals, and bait punctuation. 

%We further implement a Chromium browser extension that delivers both pre-click and post-click warnings, explanation snippets grounded in the measured features, and concise spoilers of clickbait with prompting GPT that summarize the article content. We release code, trained models, and the extension for reproducible research and practical use.

Clickbait headlines degrade the quality of online information and undermine user trust. We present a hybrid approach to clickbait detection that combines transformer-based text embeddings with linguistically motivated informativeness features. Using natural language processing techniques, we evaluate classical vectorizers, word-embedding baselines, and large language model embeddings paired with tree-based classifiers. Our best-performing model—XGBoost over embeddings augmented with 15 explicit features—achieves an F1-score of 91\%, outperforming TF–IDF, Word2Vec, GloVe, LLM prompt based classification and feature-only baselines. The proposed feature set enhances interpretability by highlighting salient linguistic cues such as second-person pronouns, superlatives, numerals, and attention-oriented punctuation, enabling transparent and well-calibrated clickbait predictions. We release code and trained models to support reproducible research.
\end{abstract}

%\includegraphics[width=\linewidth]{visual_abstract_good.png}


\section{Introduction}
\label{intro}

\begin{figure}
\centering
\includegraphics[width=\linewidth]{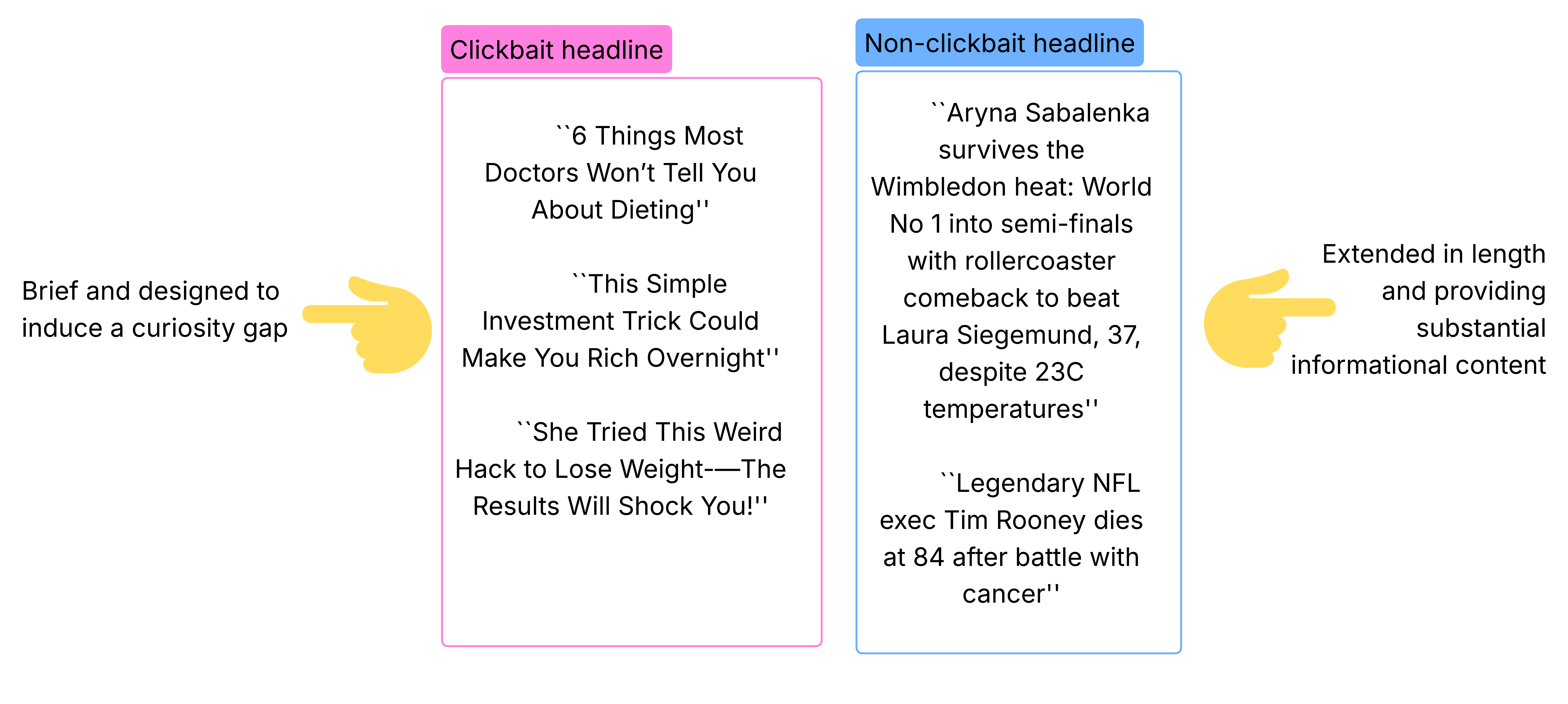}
\caption{Clickbait vs. non-clickbait headline examples.}
\label{fig:clickbait-vs-non-example}
\end{figure}

As we live in the era of digital journalism, information and news travel faster and broader than ever before \cite{berduygina2019trends}. While this rapid dissemination enables timely access to information, it also facilitates the spread of misleading or manipulative content. Disinformation—false or deceptive information shared with the intent to mislead—has therefore emerged as one of the most pressing challenges of the digital age \cite{2024under}. Clickbait represents a common and subtle manifestation of this problem (see Figure~\ref{fig:clickbait-vs-non-example}). Although clickbait headlines are not always factually false, they frequently rely on exaggerated or misleading phrasing to provoke curiosity and engagement, blurring the boundary between informative journalism and manipulation \cite{browser_extension_deep_rnn}.

Most existing research on clickbait detection formulates the task as a binary classification problem, distinguishing clickbait from non-clickbait headlines. Early work by \citep{7877426amolagarwal} demonstrated the effectiveness of deep learning approaches, employing convolutional neural networks with pre-trained word embeddings to reduce reliance on handcrafted features. \citep{bourgonje-etal-2017-clickbait} extended this line of research by modeling clickbait detection as a stance analysis problem between headlines and article bodies, thereby linking clickbait to misinformation detection. Nevertheless, their approach remained focused on categorical classification outcomes. Ensemble-based methods, such as those proposed by \citep{sisodia2019ensemble}, combined stylistic and lexical indicators to achieve strong performance and improved interpretability, but still operated within a static headline-level classification framework.

More recently, transformer-based language models, including BERT, XLNet, and RoBERTa, have significantly advanced clickbait detection performance, achieving F1-scores above 0.95 on benchmark datasets. \citep{broscoteanu-ionescu-2023-novel} introduced \textit{RoCliCo}, a manually annotated Romanian clickbait corpus, along with a contrastive \textit{Ro-BERT} model that exploits semantic similarity between headlines and article content, extending detection capabilities to low-resource languages. Despite these advances, the majority of existing approaches remain limited to binary detection, offering little insight into the informativeness of headlines or the degree to which they provide meaningful content to readers.

In this work, we retain clickbait detection as a classification task, but extend it beyond simple binary labeling. We construct a unified dataset by reorganizing and integrating multiple publicly available clickbait datasets, enabling analysis across multiple contextual dimensions rather than isolated headline classification. Building on this dataset, we propose a multi-layer detection framework. First, we introduce an explicit informativeness measure, grounded in established NLP techniques, to assess how much useful information a headline conveys. Second, we perform clickbait detection while accounting for varying degrees of bait-like characteristics, allowing for more nuanced filtering and user-facing warnings.

To evaluate our approach, we compare a wide range of methods, including traditional machine learning models, deep learning architectures, and large language models such as GPT-4, alongside transformer-based encoders including BERT and RoBERTa \cite{Liu_2023,roberta}. Our results show that explicit informativeness modeling provides strong and interpretable signals for clickbait detection, and in several settings performs competitively with—or more consistently than—approaches that rely solely on LLM prompting.

The remainder of this paper is organized as follows. Section~\ref{sec:lit-rev} reviews related work; Section~\ref{methodology} describes the proposed methodology; Section~\ref{sec:experimentation} presents the experimental setup and results; Section~\ref{discussion} discusses the findings; and Section~\ref{conclusion} concludes the paper and outlines future research directions.

\section{Literature review} \label{sec:lit-rev}
%We first consider the concept of clickbait and its defining characteristics (Section~\ref{what is clickbait}). We then review the state-of-the-art in automatic clickbait detection, tracing the evolution of methods from traditional machine learning to large language models (Section~\ref{evolution}). Finally, we examine existing tools and applications for clickbait detection, emphasizing their functionality and limitations (Section~\ref{clickbait-tools}). 

\subsection{Definition and characteristics of clickbait} \label{what is clickbait}
Although the term "clickbait" was first popularized in the mid-2000s \cite{thiel2018avoiding}, its rhetorical roots extend back much further. In Renaissance Europe, ‘bait’ already carried meanings of luring and provoking, with practices such as bear-baiting and the rhetorical tradition of copia (abundant exaggeration) functioning to entice audiences with promises of revelation. Shakespearean drama similarly ‘baited’ its audiences through hyperbolic setups that blurred the line between entertainment and knowledge. In this sense, modern clickbait continues a centuries-long tradition of using excess and curiosity as tools of persuasion \cite{hoffmann2017clickbait}.

In the contemporary era of digital journalism, however, clickbait has become institutionalized as a content strategy. As news migrated online and into social media feeds, headlines shifted from primarily informing readers to competing for their attention \cite{ferrer2018audience, shin2025emotion}. Viral sites like Upworthy in the early 2010s epitomized this shift with the ‘you’ll-never-guess-what-happened-next’ style of headline, which later spread across mainstream outlets. Today, scholars distinguish between information bait, which exploits curiosity gaps, and rage bait, which triggers outrage to drive engagement \cite{shin2025emotion}. These practices demonstrate how clickbait, once a theatrical and rhetorical device, has evolved into a systematic tool in the digital attention economy, blurring the distinction between journalism and manipulation.

\subsection{State of the art in clickbait detection} \label{evolution}
A search for clickbait detection on Google Scholar indicates that the research field was initiated by Potthast et al. \cite{potthast2016clickbait}, who introduced the first publicly available corpus of annotated Twitter posts and proposed a feature-based model evaluated with supervised classifiers, achieving an AUC of 0.79 with a random forest. In the same year, Agrawal et al. \cite{7877426amolagarwal} extended this line of research by compiling a new dataset from multiple social media platforms and employing a convolutional neural network (CNN) for classification, reaching an accuracy of 0.90 along with substantial precision and recall scores, thereby demonstrating the advantages of deep learning over traditional feature engineering approaches. Building on the Potthast dataset, Khater et al. \cite{khater} proposed a lightweight supervised approach based on 24 handcrafted features extracted from different elements of a post (post text, title, and article body), achieving an F1-score of 79\% and an AUC of 70\%, thereby showing that exploiting multiple post components can be practical even with minimal features. According to a recent review by Raj et al. \cite{rajetall}, subsequent studies have continued to expand dataset development. From 2021 onward, the field began to systematically include low-resource languages such as Turkish and Indonesian, marking a departure from the earlier English-dominated focus. Raj et al. \cite{rajetall} further emphasize that while diverse methods and datasets have emerged, much of the work remains centered on benchmarking classification performance, with limited attention to how these approaches can be operationalized in practical tools and applications.

\subsection{Research gap} \label{gaps}
Despite significant advances in clickbait detection, with recent models reporting accuracy levels above 90\% \cite{jeevashri2025hybrid, 7877426amolagarwal}, the task is still predominantly addressed as a black-box binary classification problem. Although such approaches achieve strong predictive performance, they provide limited insight into the linguistic and stylistic factors that contribute to clickbait characteristics.

Previous studies have identified a variety of relevant signals, including morphological patterns \cite{vincze-szabo-2020-automatic}, readability measures \cite{jeevashri2025hybrid}, and semantic features \cite{pujahari2021clickbait}. However, these cues are often incorporated implicitly within complex model architectures and are rarely exposed in an interpretable manner. Consequently, the connection between model predictions and human-understandable indicators of manipulation remains insufficiently explored.

This gap highlights the need for approaches that not only achieve high detection accuracy but also offer greater transparency regarding the features and representations driving classification decisions. In particular, combining modern embedding techniques with explicit, linguistically motivated features remains an underexplored direction in clickbait detection research.

% ================== MAIN TEXT ==================
\section{The proposed model}\label{methodology}

\subsection{Task definition}\label{task-definition}
We address the task of clickbait detection. Given a headline (and, where applicable, its associated article), the goal is to determine whether the headline exhibits clickbait characteristics. Our approach represents text using multiple embedding strategies, ranging from traditional TF-IDF and word-embedding models to transformer-based representations, which are subsequently used as inputs to supervised classification models. The experimental setup is inspired by prior work and benchmarks from the Clickbait Challenge 2017.

\subsection{Constructed dataset}
\label{datasets}

\subsubsection{Data sources}
\label{data_sources}

We utilize four publicly available English datasets that cover clickbait detection and spoiling. The first two, \textbf{Kaggle-1}\cite{kaggle_1_source}\footnote{\url{https://kaggle.com/datasets/amananandrai/clickbait-dataset}} and \textbf{Kaggle-2}\cite{kaggle_2_source}\footnote{\url{https://kaggle.com/datasets/vikassingh1996/news-clickbait-dataset}}, contain 32{,}000 and 21{,}029 annotated news articles, respectively. Kaggle-1 headlines originate from outlets such as BuzzFeed, Upworthy, and ViralNova, while Kaggle-2 was released as part of a Kaggle competition. 
We restrict Kaggle-2 to the \texttt{train2.csv} split, since \texttt{train1.csv} overlaps with Kaggle-1. The third dataset, \textbf{Clickbait Challenge 2017}\cite{clickbait_challenge_2017}\footnote{\url{https://webis.de/events/clickbait-challenge/shared-task.html}}, consists of 38{,}830 Twitter posts from 27 US news outlets. Each post was annotated on a four-point scale (\(0.0, 0.33, 0.66, 1.0\)) by five crowd workers. In our experiments, we use the \texttt{postText} field as a proxy for the headline and the median annotation score (\texttt{truthMedian}) as the label. Finally, the \textbf{SemEval Clickbait Spoiling dataset}\cite{pan_webis_spoiling,hagen2022clickbaitspoilingquestionanswering}\footnote{\url{https://pan.webis.de/semeval23/pan23-web/clickbait-challenge.html}} 
contains about 4{,}000 posts collected from social media accounts dedicated to exposing clickbait (e.g., \texttt{r/savedyouaclick}, \texttt{@SavedYouAClick}). 
Some entries also originate from the Clickbait Challenge 2017, featuring manually generated spoilers. The dataset provides both extractive and abstractive spoilers, enabling experiments beyond binary classification. Table~\ref{tab:datasets} summarizes dataset statistics and collection methods.

\begin{table}[!ht]
\centering
\scriptsize
\caption{Summary of datasets for clickbait detection and spoiling, where Instances denotes the number of rows in each dataset}
\begin{tabular}{p{2.3cm}ccp{3.2cm}p{3.3cm}}
\toprule
\textbf{Name} & \textbf{Instances} & \textbf{Task} & \textbf{Features} & \textbf{Collection} \\
\midrule
Kaggle-1 & 32{,}000 & Detection & Title, binary label & News websites (method undocumented) \\
Kaggle-2 & 21{,}029 & Detection & Title, body, binary label & Kaggle competition \\
Clickbait Challenge 2017 & 38{,}830 & Detection & Title (\texttt{postText}), score (0.0–1.0) & Twitter; crowdsourced labels \\
SemEval Spoiling & $\sim$4{,}000 & Spoiling & Title, body, extractive + abstractive spoilers & Social media spoiling accounts \\
\bottomrule
\end{tabular}
\label{tab:datasets}
\end{table}

\subsubsection{Our derived datasets}
\label{derived_datasets}

We prepared three task-specific working sets with light harmonization (English-only, drop empty titles and duplicates, binary label harmonization across sources—Kaggle datasets are already binary; Clickbait Challenge 2017 (CC17)’s graded annotations are thresholded at $0.5$ and normalized to $0$ for non-clickbait and $1$ for clickbait; stratified $80/10/10$ splits; see~\ref{app:label_harmonization} for details). Datasets were constructed, following standard practices in clickbait detection research \cite{chakraborty2016stop,potthast2018clickbaitchallenge2017regression}.

\begin{itemize}
  \item \textbf{Title-only detection.} Concatenation of Kaggle-1, Kaggle-2, and Clickbait Challenge 2017 (CC17). From CC17 we use \texttt{targetTitle} (fallback: first \texttt{postText}) and binarize labels via $y=\mathbb{1}[\texttt{truthMedian}\ge 0.5]$ (we also checked $0.33/0.66$). Fields are harmonized to \texttt{title}/\texttt{label}. Besides the full concat, we train on a balanced sample ($20{,}000$/$20{,}000$). Splits are stratified $80/10/10$.

  \item \textbf{Title+body detection.} Subset of CC17 with both \texttt{targetTitle} and \texttt{targetParagraphs} present; the label uses the same mapping $y=\mathbb{1}[\texttt{truthMedian}\ge 0.5]$. Same cleaning and stratified $80/10/10$ split.

  \item \textbf{Spoiling.} SemEval Clickbait Spoiling dataset using \texttt{targetTitle}, \texttt{targetParagraphs}, \texttt{humanSpoiler} (abstractive), \texttt{spoiler} (extractive), and \texttt{tags}. For abstractive training, we drop rows with missing \texttt{humanSpoiler}. Splits are stratified $80/10/10$ (by \texttt{tags}).
\end{itemize}

\subsection{Our approach}\label{sub:main-conception}

\begin{figure}
    \centering
    \includegraphics[width=\linewidth]{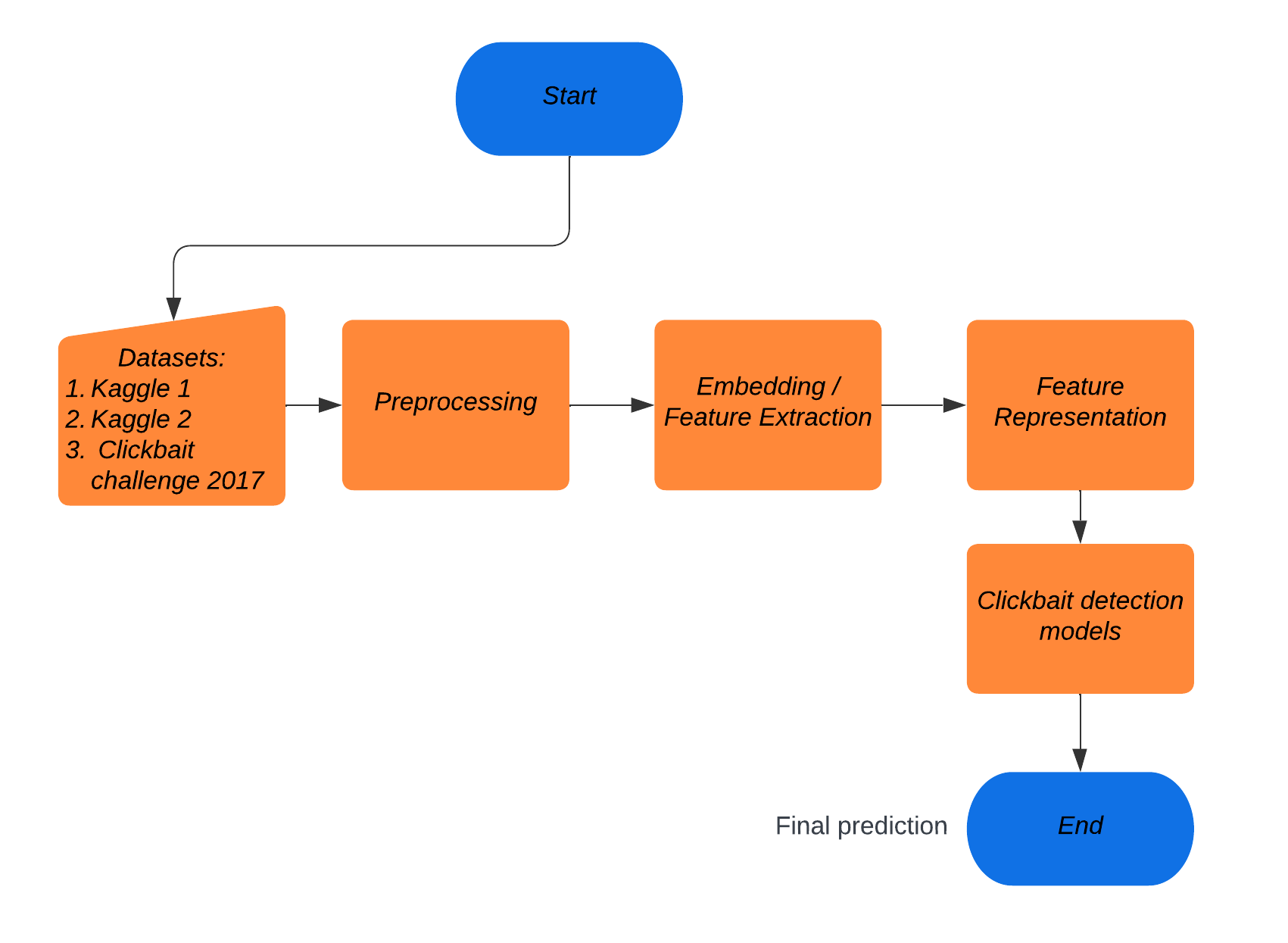}
    \caption{Overview of the experimental workflow. After dataset collection and preprocessing, embeddings and features are generated using TF-IDF, Word2Vec, TF-IDF-weighted Word2Vec, GloVe, and transformer-based models (BERT and OpenAI). These representations are then fed into clickbait detection models based on Random Forest and XGBoost to obtain the final prediction.}
    \label{fig:workflow-main-concept}
\end{figure}

The proposed approach (Figure~\ref{fig:workflow-main-concept}) follows an experimental pipeline for clickbait detection. The methodology integrates multiple datasets, applies uniform preprocessing and balancing procedures, and constructs textual representations using both embedding-based methods and explicit linguistic features. These representations are then used to train and evaluate supervised classification models for clickbait detection. A consistent feature extraction and evaluation setup is maintained across all models to ensure comparability of results. %Detailed preprocessing steps, feature definitions, and training configurations are provided in ~\ref{app:offline}.

\subsection{Informativeness measures}\label{inf_mes}
We treat informativeness as a construct reflected by quantifiable stylistic and linguistic properties—readability, lexical diversity, sentiment, and usage patterns—rather than something directly measurable \citep{nlp_in_action}. In practice, we compute a compact set of representative measures in the main text (Table~\ref{tab:core_measures}) and rely on an expanded battery of 25 measures, complete definitions, and equations in~\ref{app:inf_measures_full}. 

\begin{table}[t]
\centering
\small
\caption{Representative informativeness measures (complete list and equations in~\ref{app:inf_measures_full}).}
\label{tab:core_measures}
\begin{tabular}{lp{9cm}}
\toprule
Measure & Description \\
\midrule
Word Count & Total number of words in the text \\
Common Words Ratio & Proportion of stop words to total words \\
Capital Letters Ratio & Share of uppercase characters \\
Superlatives Ratio & Frequency of superlative adjectives or adverbs \\
FRES & Flesch Reading-Ease score; higher indicates easier readability \\
Polarity & Sentiment polarity of the text \\
Similarity & Cosine similarity between title and article body \\
\bottomrule
\end{tabular}
\end{table}

\subsection{Clickbait detection}\label{sec:clickbait_detection_methodology}
Our detector evaluates multiple text representations and classifiers, ultimately adopting a hybrid approach that concatenates manually engineered informativeness measures with transformer-based embeddings, followed by XGBoost classification. This design leverages explicit stylistic cues in conjunction with deep semantic signs. We report headline-level performance using standard binary metrics and select the model based on $F_1$ as the primary criterion. Comprehensive feature definitions, the complete set of representations (TF–IDF, Word2Vec, GloVe, transformer embeddings), model families (tree ensembles and transformer-based classifiers), hyperparameters, and ablations are provided in~\ref{app:det_ext}. %Precise metric formulas and additional plots appear in Appendix~\ref{app:spoil_ext}.

\section{Experimentation}
\label{sec:experimentation}

To rigorously evaluate the proposed clickbait detection approach, we designed a multifaceted experimental protocol. The evaluation aims to: (1) benchmark the proposed hybrid model against a comprehensive set of baselines; (2) conduct a formal ablation study to isolate the contribution of the engineered informativeness features; (3) analyze model robustness and behavior; and (4) assess the interpretability benefits introduced by explicit feature representations.

\subsection{Experimental environment setting}
\label{xperimental-setup}

\paragraph{Environment}
All data preparation, feature extraction, and model training were conducted on a local CPU-only workstation with 32\,GB RAM. The software stack consisted of Python~3.11, scikit-learn, XGBoost, spaCy, and the Hugging Face Transformers library. Transformer-based text embeddings were obtained using OpenAI API endpoints. To ensure reproducibility, all datasets were split into stratified training (80\%), validation (10\%), and test (10\%) sets, with random seeds fixed to 42.

\paragraph{Evaluation Metrics}
For the clickbait detection task, we report Accuracy, Precision, Recall, F1-score, and ROC AUC. Model selection is guided by the F1-score on the validation set, and all final results are reported on the held-out test set.

\subsection{Baselines and models}
\label{baseline and models}

To contextualize performance, we compare the proposed hybrid model against a diverse set of baselines representing the evolution of clickbait detection techniques. Initial benchmarks are established using traditional lexical models, pairing a TF-IDF vectorizer with a centroid-based classifier and with XGBoost. We further evaluate static embedding models (Word2Vec and GloVe) to assess non-contextual semantic representations. Transformer-based baselines include a publicly available fine-tuned RoBERTa detector \cite{roberta_hugging_face} and a model using high-dimensional OpenAI text embeddings paired with an XGBoost classifier.

To directly test our central hypothesis, two additional models are evaluated: a feature-only model trained exclusively on the 15 handcrafted informativeness features, and the proposed hybrid model, which combines these features with reduced-dimensional OpenAI embeddings to jointly capture stylistic and deep semantic signals.

\subsection{Comparative analysis}

Table~\ref{tab:comparative_analysis} presents the performance of all models on the title-only clickbait detection task. The results demonstrate a clear progression in performance as model expressiveness increases. While traditional lexical representations provide strong baselines, transformer-based embeddings substantially improve predictive accuracy. The proposed hybrid model consistently outperforms all baselines across evaluation metrics. We additionally report zero-shot and few-shot performance of large language models as prompt-based baselines. The exact prompt templates used for zero-shot and few-shot inference, including output constraints and example selection, are provided in~\ref{app:llm_prompts}.

\begin{table}[!ht]
\centering
\small{
\caption{Performance comparison of clickbait detection models on the held-out test set.}
\label{tab:comparative_analysis}
\begin{tabular}{p{2.7cm} lccccc}
\toprule
\textbf{Representation} & \textbf{Classifier} & \textbf{Acc.} & \textbf{Prec.} & \textbf{Rec.} & \textbf{F1} & \textbf{AUC} \\
\midrule
\multicolumn{7}{l}{\textit{Traditional Baselines}} \\
TF-IDF (raw)        & Centroid        & 0.786 & 0.884 & 0.659 & 0.755 & --    \\
TF-IDF (raw)        & Random Forest   & 0.832 & 0.842 & 0.816 & 0.829 & 0.906 \\
Word2Vec (Google)   & XGBoost         & 0.832 & 0.840 & 0.818 & 0.829 & 0.908 \\
GloVe (100d)        & XGBoost         & 0.832 & 0.840 & 0.818 & 0.829 & 0.908 \\
\midrule
\multicolumn{7}{l}{\textit{Transformer Baselines}} \\
RoBERTa (HF finetuned) & --           & 0.832 & 0.795 & 0.894 & 0.842 & 0.910 \\
OpenAI emb (3072d)     & XGBoost      & 0.866 & 0.877 & 0.851 & 0.864 & 0.938 \\
\midrule
\multicolumn{7}{l}{\textit{Prompt-based LLM Baselines}} \\
GPT-4o-mini     & Zero-shot & 0.72 & 0.76 & 0.72 & 0.71 & -- \\
GPT-4o-mini     & Few-shot  & 0.82 & 0.82 & 0.82 & 0.82 & -- \\
Gemini-3 Flash  & One-shot  & 0.83 & 0.84 & 0.83 & 0.83 & -- \\
Gemini-3 Flash  & Few-shot  & 0.84 & 0.84 & 0.84 & 0.84 & -- \\
\midrule
\multicolumn{7}{l}{\textit{Proposed Hybrid Model}} \\
\textbf{OpenAI emb (1000d) + Features} & \textbf{XGBoost} & \textbf{0.910} & \textbf{0.922} & \textbf{0.896} & \textbf{0.909} & \textbf{0.969} \\
\bottomrule
\end{tabular}
}
\end{table}

\subsection{Ablation study: Impact of informativeness features}

To isolate the contribution of the handcrafted informativeness features, we conducted a formal ablation study comparing the hybrid model against its constituent components: an embeddings-only model and a features-only model. Results are reported in Table~\ref{tab:ablation_study}.

\begin{table}[!ht]
\centering
\caption{Ablation study evaluating the contribution of the 15 handcrafted informativeness features.}
\label{tab:ablation_study}
\begin{tabular}{p{4.6cm} l c p{3cm}}
\toprule
\textbf{Model inputs} & \textbf{Classifier} & \textbf{F1-score} & \textbf{Improvement ($\Delta$) vs. Emb-only} \\
\midrule
Features only               & Random Forest & 0.821 & -- \\
OpenAI emb (3072d)           & XGBoost      & 0.864 & -- \\
\midrule
\textbf{OpenAI emb (1000d) + Features} & \textbf{XGBoost} & \textbf{0.909} & \textbf{+4.5 points} \\
\bottomrule
\end{tabular}
\end{table}

The results demonstrate that while transformer-based embeddings provide a strong baseline, the handcrafted features alone also capture substantial predictive signal. Most importantly, their combination yields a significant improvement, confirming that the features encode complementary stylistic and structural cues not fully captured by dense semantic embeddings.

\section{Discussion} \label{discussion}

As shown in Table X, the proposed hybrid model combining OpenAI embeddings (1000d) with explicit linguistic features and an XGBoost classifier achieves the strongest overall performance, with an accuracy of 0.910, precision of 0.922, recall of 0.896, F1-score of 0.909, and an AUC of 0.969. These results represent a substantial improvement over all baseline approaches, including transformer-based and prompt-based large language models.

Compared to the strongest embedding-only baseline using OpenAI embeddings (3072d), which achieves an F1-score of 0.864 and an AUC of 0.938, the hybrid model improves F1 by 4.5 percentage points and AUC by 3.1 points. This gain indicates that the added linguistic features contribute information that is not redundant with high-dimensional contextual embeddings.

Prompt-based large language models, such as Gemini-3 Flash (few-shot) and GPT-4o-mini (few-shot), achieve F1-scores of 0.84 and 0.82, respectively, which are notably lower than the hybrid approach despite being trained on large-scale corpora. This suggests that scale and general contextual knowledge alone do not ensure consistent sensitivity to task-specific linguistic cues relevant to clickbait detection.

Overall, these findings indicate that explicitly modeling linguistic and stylistic features provides added value even when strong pretrained representations are available. In particular, the results suggest that contextual understanding acquired through large-scale pretraining benefits from task-specific grounding when the target phenomenon depends on pragmatic language use rather than semantic content alone.

\section{Conclusion and future work}
\label{conclusion}

\section*{Declaration of generative AI and AI-assisted technologies in the manuscript preparation process}
\noindent During the preparation of this work, we used Grammarly and ChatGPT to enhance language clarity and readability, while carefully reviewing and editing all content. The authors take full responsibility for the final version of the manuscript.

\clearpage
\appendix

\section{Informativeness measures (full)}\label{app:inf_measures_full}

\subsection{Complete list and definitions}
Table~\ref{table:informativeness_measures} lists the 25 measures used.

\begin{table}[!ht]
\caption{Informativeness measures used}
\label{table:informativeness_measures}
\begin{tabular}{r| p{3.5cm}  p{10cm}}
\toprule
\textbf{No.} & \textbf{Name} & \textbf{Description} \\
\midrule
1 & Character count & Number of characters, including whitespaces \\
2 & Word count & Number of words \\
3 & Mean word length & Average length of words \\
4 & Common words ratio & Percentage of stop words \\
5 & Capital letters ratio & Percentage of uppercase characters \\
6 & Capital words count & Number of fully capitalized words (excluding one-letter words) \\
7 & Punctuation ratio & Percentage of punctuation/special characters \\
8 & Bait punctuation count & Punctuation related to bait titles: ! " ( ? \# \\
9 & Non-bait punctuation count & Punctuation related to non-bait titles: \$ \% \& , . ; : - / \\
10 & Numbers count & Count of numbers (not digits) \\
11 & Pronouns & Number of pronouns (e.g., ``she'', ``his'', ``you'') \\
12 & 2nd person pronouns & Count of ``you'', ``your'', ``yours'', ``yourself'', ``yourselves'' \\
13 & Superlatives ratio & Share of adjectives/adverbs in superlative form \\
14 & Speculatives usage & Count of words like ``may'', ``might'', ``possibly'' \\
15 & Bait phrases usage & Count of clickbait lexemes (e.g., ``tweets'', ``hilarious'') \\
16 & Similarity score & Title–body cosine similarity (Word2Vec) \\
17 & Polarity score & Sentiment polarity (TextBlob) \\
18 & Subjectivity score & Subjectivity (TextBlob) \\
19 & TTR & Type–token ratio \\
20 & CTTR & $CTTR=\frac{t}{\sqrt{2n}}$ where $t$ types, $n$ tokens \\
21 & Maas Index & $a^2=\frac{\log n - \log t}{(\log n)^2}$ \\
22 & HD-D & Hypergeometric diversity (average TTR on 42‑word samples) \\
23 & FRES & $FRES=206.835-1.015\frac{words}{sentences}-84.6\frac{syllables}{words}$ \\
24 & FKGL & $FKGL=0.39\frac{words}{sentences}+11.8\frac{syllables}{words}-15.59$ \\
25 & ARI & $ARI=4.71\frac{characters}{words}+0.5\frac{words}{sentences}-21.43$ \\
\bottomrule
\end{tabular}
\end{table}

\subsection{Baitness score derivation}\label{app:baitness_formula}
We define the composite score as:
$$
\text{Baitness}=\frac{\text{EyeCatch}+\text{Curiosity}+\text{Sentiment}+\text{EaseOfText}}{4}.
$$
Subcomponents:
$$
\text{EyeCatch}=\min\Bigl(1,\max\bigl(0,\frac{\text{BaitPunct} + 3\cdot\text{CapitalsRatio} + \text{Numbers}}{3}\bigr)\Bigr),
$$
$$
\text{Curiosity}=\sqrt{\min\Bigl(1,\max\bigl(0,\frac{\text{2ndPronouns} + 2\cdot\text{Superlatives} + \text{Speculatives} + \text{BaitWords}}{4}\bigr)\Bigr)},
$$
$$
\text{Sentiment}=\sqrt{|\text{Polarity}|\cdot \text{Subjectivity}},
$$
$$
\text{EaseOfText}=\frac{\min(1,\text{FRES}/100)+\min\bigl(1,1.5\cdot\text{CommonWordsRatio}\bigr)}{2}.
$$

\section{Clickbait detection: extended experiments}\label{app:det_ext}

\subsection{Feature representation methods and settings}
\begin{itemize}
\item TF–IDF: \texttt{ngram\_range}$(1,2)$; raw vs.\ cleaned variants as described above.
\item Word2Vec: mean‑pooled embeddings; Google News (300d) and in‑house CBOW (\texttt{window}$=5$, \texttt{min\_count}$=2$, \texttt{epochs}$=10$).
\item GloVe: pre‑trained 100d.
\item OpenAI embeddings: \texttt{text-embedding-3-large}\\ with dimensions $\{3072,1000,100,30\}$; L2 normalization.
\end{itemize}

\subsection{Classification models}
Random Forest and XGBoost on each representation; public RoBERTa detector (HuggingFace) as a strong baseline; direct prompting (GPT‑4o‑mini/ GPT‑3.5‑turbo) with constrained outputs and exemplar prompt.

\subsection{Hybrid approach}
Early fusion by concatenating the 15‑measure vector with the sentence embedding. XGBoost was chosen as the final model due to its top $F_1$ score and calibration. Threshold tuning is considered, but the default value of $0.5$ is used for consistency.

\subsection{Evaluation metrics for clickbait detection}
We report Accuracy, Precision, Recall, $F_1$, and ROC–AUC.
$$
\text{Accuracy}=\frac{TP+TN}{TP+TN+FP+FN},$$ \\
$$\text{Precision}=\frac{TP}{TP+FP},\quad 
\text{Recall}=\frac{TP}{TP+FN},$$\quad\\
$$F_1=2\cdot\frac{\text{Precision}\cdot\text{Recall}}{\text{Precision}+\text{Recall}}.
$$

\section{LLM prompts}\label{app:llm_prompts}

\subsection{Zero-shot classification prompt}

\begin{lstlisting}[
    caption={Few-shot LLM prompt for clickbait detection},
    label={lst:fewshot_prompt},
    breaklines=true,
    breakatwhitespace=true,
    columns=fullflexible,
    basicstyle=\ttfamily\footnotesize,
    xleftmargin=1em
]
You are a strict classifier.

Task:
Determine whether a news article title is clickbait.

Definition:
Clickbait is a title that is intentionally sensational, misleading, or emotionally manipulative
in order to attract clicks, often by exaggerating, omitting key facts, or creating a curiosity gap.

Labels:
0 - non-clickbait
1 - clickbait

Rules:
- Base your decision only on the title.
- Do not provide explanations or justifications.
- Output only a single digit: 0 or 1.
\end{lstlisting}

\subsection{Few-shot classification prompt}

\begin{lstlisting}[
    caption={Few-shot LLM prompt for clickbait detection},
    label={lst:fewshot_prompt},
    breaklines=true,
    breakatwhitespace=true,
    columns=fullflexible,
    basicstyle=\ttfamily\footnotesize,
    xleftmargin=1em
]
You are a strict classifier.

Task:
Determine whether a news article title is clickbait.

Definition:
Clickbait is a title that is intentionally sensational, misleading, or emotionally manipulative
in order to attract clicks, often by exaggerating, omitting key facts, or creating a curiosity gap.

Labels:
0 - non-clickbait
1 - clickbait

Rules:
- Base your decision only on the title.
- Do not provide explanations or justifications.
- Output only a single digit: 0 or 1.

Clickbait examples:
<CLICKBAIT_EXAMPLES>

Non-clickbait examples:
<NON_CLICKBAIT_EXAMPLES>
\end{lstlisting}

\end{document}